\documentclass{article}

\usepackage{PRIMEarxiv}

\usepackage[utf8]{inputenc} 
\usepackage[T1]{fontenc}    
\usepackage{hyperref}       
\usepackage{url}            
\usepackage{booktabs}       
\usepackage{amsfonts}       
\usepackage{nicefrac}       
\usepackage{microtype}      
\usepackage{lipsum}
\usepackage{fancyhdr}       
\usepackage{graphicx}   
\usepackage{multicol}
\usepackage{multirow}
\usepackage{adjustbox}
\graphicspath{{media/}}     

\title{Multimodal Stress Detection Using Facial Landmarks and Biometric Signals
\thanks{Corresponding Author: 
\textbf{Raju Gottumukkala}, \underline{Raju.Gottumukkala}@\underline{louisiana.edu}} 
}

\author{
  Majid Hosseini, Morteza Bodaghi, Ravi Teja Bhupatiraju, Anthony Maida, Raju Gottumukkala*\\
  University of Louisiana at Lafayette\\
  USA\\
  \texttt{mjhoseiny@gmail.com, Bodaghi.Morteza@gmail.com, raviteja@louisiana.edu}\\ \texttt{Maida@louisiana.edu, raju@louisiana.edu*} \\
   \And
}

\begin{document}
\maketitle

\begin{abstract}
The development of various sensing technologies is improving measurements of stress and the well-being of individuals. Although progress has been made with single signal modalities like wearables and facial emotion recognition, integrating multiple modalities provides a more comprehensive understanding of stress, given that stress manifests differently across different people. Multi-modal learning aims to capitalize on the strength of each modality rather than relying on a single signal. Given the complexity of processing and integrating high-dimensional data from limited subjects, more research is needed. Numerous research efforts have been focused on fusing stress and emotion signals at an early stage, e.g., feature-level fusion using basic machine learning methods and 1D-CNN Methods. This paper proposes a multi-modal learning approach for stress detection that integrates facial landmarks and biometric signals. We test this multi-modal integration with various early-fusion and late-fusion techniques to integrate the 1D-CNN model from biometric signals and 2-D CNN using facial landmarks. We evaluate these architectures using a rigorous test of models' generalizability using the leave-one-subject-out mechanism, i.e., all samples related to a single subject are left out to train the model. Our findings show that late-fusion achieved 94.39\% accuracy, and early-fusion surpassed it with a 98.38\% accuracy rate. This research contributes valuable insights into enhancing stress detection through a multi-modal approach. The proposed research offers important knowledge in improving stress detection using a multi-modal approach. 
\end{abstract}

\section{Introduction}
Stress has gained importance in recent years as people are placing more emphasis on holistic well-being, i.e., physical health, mental health, emotional and social well-being. Moreover, prolonged exposure to stress can in turn, affect physical and mental health, leading to conditions such as heart disease, depression, and anxiety \cite{cohen2007psychological}.  Recent research has shown the potential of employing computer vision and biometric data for stress detection \cite{alberdi2016towards}. Biometric indicators such as Heart Rate (HR), Electrodermal Activity (EDA), and Skin Temperature (ST/TEMP)  have the capacity to reflect an individual's physiological reactions to stress, making them reliable objective measures of stress levels. Conversely, facial landmarks represent crucial points on the face that provide valuable insights into facial expressions, enabling inference of emotional states and stress. Conventional stress detection methods mostly focus on a single modality, such as physiological signals or facial expressions. It is well known that Stress is a complex interplay of psychological and physiological responses to external or internal stimuli, making it difficult to predict accurately. One of the key challenges of predicting stress is that it is transient, meaning it can change quickly and is influenced by a variety of factors. These factors include environmental stimuli, prior conditioning or past experience with stress, emotional states, and external factors such as work demands, social interactions, and life events. These factors can significantly impact stress levels and vary widely between individuals. Stress manifests differently in different people, leading to individual differences in stress responses \cite{berger2015stress}. This makes it difficult to develop a single, accurate method for stress prediction. Researchers are investigating new approaches to stress prediction that use multiple data sources such as heart rate, temperature, and facial expressions \cite{ hosseini2022development, islam2023personalized, bobade2020stress, zhang2022real}. The idea of multimodal stress prediction is inspired by the human senses, which work together to provide us with a comprehensive understanding of the world around us.

It is also important to differentiate between "multivariate" and "multimodal" learning for neural networks. Multivariate networks focus on processing multiple variables or predictors but often the same kind of data. For example, it might analyze time series data signals such as ECG, temperature, and heart rate, all intended for stress detection. In contrast, multimodal neural networks are designed to integrate information from different types of data sources or modalities. For example, a multimodal network is one that processes voice and image data simultaneously, determining how one might be related to one another. Multimodal stress prediction models combine data from multiple sensors to learn complex patterns that are associated with stress.

This paper presents a novel multimodal approach for stress detection, combining facial landmarks analysis with biometric signals comparing early and late fusion techniques. The overarching theme of stress detection is pre-established, and various fusion methodologies, both late and early, have been explored previously \cite{zhang2022real,bobade2020stress,montesinos2019multi,kuttala2023multimodal}. Earlier research typically treats both modalities as time series signals, modeling each with a 1D CNN. In our approach, we model facial landmarks using a 2D CNN. The integration of 2D CNN and 1D CNN models for stress detection has not seen studied before. We study the performance of this against a wide range of both late and early fusion techniques that employ facial landmarks and biometric feedbacl. We also analyze the effects of various signal components and model architectures, including 2079 video and biometric signal features, EDA tonic and phasic components, and the performance of six deep learning models with different feature selection methods, along with eleven multimodal models using these features.

We first discuss related work in the area of stress detection, focusing on physiological signals from wearables and computer vision, as well as multi-modal stress detection approaches. We provide a comprehensive summary of biomedical features documented in the literature. Next, we offer an in-depth overview of the dataset and elaborate on the methodologies employed to extract various features from physiological and computer vision datasets. Following this discussion, we present early and late-fusion models and their architectures. Finally, we evaluate and discuss the outcomes of the study, weighing various model architectures against both accuracy and computational cost.

\section{Related Work}
Researchers have studied stress detection through the use of wearable devices and facial expressions. Sioni et al. \cite{sioni2015stress} discuss how physiological signals such as HR and EDA can detect stress. Alberdi et al. \cite{alberdi2016towards} studied automatic, continuous, and unobtrusive early stress detection methods for office workers. They found that ECG, especially using HRV features, and EDA are the most accurate physiological signals for recognizing stress. It is desirable to have better accuracy when incorporating data from multiple modalities, such as behavioral responses. This approach would lead to the development of a less intrusive and widely applicable monitoring system, significantly enhancing practicality. Well-engineered multimodal methods should offer reliable models through diverse signal information and enhanced robustness against noise.
We review previous research in the fields of physiological signals, computer vision, and multi-modal signals.

\begin{table*}
\caption{Summarize of Reviewed Paper}
\resizebox{\textwidth}{!}{%
\begin{tabular}{|p{3.8cm}|p{7cm}|p{3.2cm}|}
\hline
\textbf{Author}                                                              & \textbf{Signals}                                                                     & \textbf{\begin{tabular}[c]{@{}l@{}}Classification\\    Performance \end{tabular}}                                                                                 \\ \hline

Schmidt, Philip, et al. \cite{schmidt2018introducing}        & BVP, ECG, EDA, EMG, RR, TEMP, and ACC.             &
\begin{tabular}[c]{@{}l@{}}Acc 80\% (3-class)\\    93\% (2-class) \end{tabular} \\ \hline

Koldijk, et al. \cite{koldijk2016detecting}                  & Facial Expressions, Posture, Physiology.                       & Acc 90\% (2-class)                                                                                         \\ \hline
Giannakakis, et al. \cite{giannakakis2017stress}   & Facial Cues and camera-based HR.   & Acc 91.68\% (3 -class)                                                                                      \\ \hline

Padmaja, B., et al. \cite{padmaja2018machine}               & FITBIT wearable device data                                                    & Acc 62\% (3-class)                                                                                         \\ \hline

Shu, Lin, et al. \cite{shu2020wearable}                     & HR                                                                                   & Acc 84\% (3-class)                                                                                       \\ \hline
Hsu, Yu Liang, et al. \cite{hsu2020automatic}               & ECG                                                                                  & Acc 82.78\% (2-class)                                                                                                                                                                            \\ \hline
Sharma,  et al. \cite{sharma2020automated}                  & EEG signals                                                                          & Acc 82.01\% (4-class)                                                                                      \\ \hline
Zhu, Lili, et al.   \cite{zhu2022feasibility}               & EDA                                                                                  & Acc 85.7\% (2-class)                                                                                       \\ \hline
Zhu, Lili, et al. \cite{zhu2022multimodal}                  & EDA, ECG, and PPG                                             & Acc 86.4\% (2-class)                       \\ \hline
Jeon, Taejae, et al. \cite{jeon2020stress}                  & Facial Landmarks                                                                     & Acc 64.63\% (3-class)                                                                                      \\ \hline
Zhang, Huijun, et al. \cite{zhang2020video}                 & Facial Expressions                                                                   & Acc 85.42\% (2-class)                                                                                      \\ \hline
Cardone, Daniela, et al.   \cite{cardone2020driver}         & Thermal Infrared Imaging                                                             & \begin{tabular}[c]{@{}l@{}}AUC 80\% (2-class)\end{tabular} \\ \hline
Kopaczka, Marcin, et al.   \cite{kopaczka2019modular}       & Facial Expressions                                                                   & Acc 65.75\% (4-class)                                                                                      \\ \hline
Seo, Wonju, et al. \cite{seo2022deep}                       & ECG, RR, and Facial Landmarks                      & Acc 73.3\% (2-class)                                                                                     \\ \hline

Bobade, Pramod et al.   \cite{bobade2020stress}             & ACC, ECG, BVP, TEMP, RR, EMG and EDA                                                &
\begin{tabular}[c]{@{}l@{}}Acc 84.32 \% (3-class)\\    95.21\% (2-class) \end{tabular} \\ \hline

Walambe, Rahee, et al.   \cite{walambe2021employing}        & Facial Expressions, Posture, HR ,Computer Interaction               & Acc 96.09\% (2-class)                                                                                      \\ \hline
Naegelin, Mara, et al.   \cite{naegelin2023interpretable}   & HR and Behavioral data                                                               & F1 score 77.5\% (3-class)                                                                               \\ \hline
Kuttala, Radhika et al.   \cite{kuttala2023multimodal}      & EDA and ECG                                                                          & Acc 97.6\% (2-class)                                                                                     \\ \hline
Hosseini, Majid, et al.   \cite{hosseini2022empathicschool} & HR, TEMP, EDA, ACC and Facial Expression                                                      & Prec 85.04 (3-class)                                                                               \\ \hline

\end{tabular}
}
\label{tab:1}
\end{table*}

\subsection{Physiological Signals}
Schmidt et al. \cite{schmidt2018introducing} introduced the WESAD dataset consisting of BVP, ECG, EDA, EMG, RR, ST, and three-axis acceleration (ACC) to detect neutral, stress, and amusement using self-reports validation surveys. The benchmark achieved classification accuracies up to 80\% for three classes (baseline, stress, amusement) and up to 93\% for binary (stress, non-stress) classification. Padmaja, B. et al. \cite{padmaja2018machine} propose a method for detecting stress levels using data from a physical activity tracker device (Fitbit), using Logistic Regression to evaluate the impact of individual stressors. They achieved 62\% accuracy for three stress levels. Hsu, Yu Liang, et al. \cite{hsu2020automatic} introduced an automatic algorithm for emotion recognition based on ECG signals. The classification rates for positive/negative valence, high/low arousal, and four emotion classification tasks were 82.78\%, 72.91\%, and 61.52\%, respectively. Shu, Lin, et al. \cite{shu2020wearable} achieved 84\% accuracy for emotion recognition based on HR data from a wearable smart bracelet using the Gradient Boosted Decision Trees (GBDT) model. Sharma et al. \cite{sharma2020automated} achieved an accuracy of 82.01\% with four-labeled emotion classes using 10-fold cross-validation on online recognition of human emotions using EEG signals on the DEAP dataset \cite{koelstra2011deap}. Zhu, Lili, et al. \cite{zhu2022feasibility} examined the potential of utilizing EDA from two publicly available datasets collected from research-grade wearable devices. The experimental results demonstrated that Random Forests(RF) achieve an 85.7\% accuracy rate in classifying stress from non-stress states. Zhu, Lili, et al. 
\cite{zhu2022multimodal} used signals of EDA, ECG, and photoplethysmography (PPG) from smartwatches. They investigated machine learning methods for two-level stress classification. The results indicate that EDA, particularly when used with Stacking Ensemble Learning, achieves the highest accuracy of 86.4\% compared to other signals and combinations. Kuttala, Radhika, et al. \cite{kuttala2023multimodal} used EDA and ECG signals for the hierarchical feature fusion stress detection model and achieved 97.6\% accuracy for binary classification.

\subsection{Imaging/Video Signals}
Giannakakis, Giorgos, et al. \cite{giannakakis2017stress} introduces a framework for detecting stress and anxiety emotional states using facial cues captured in video recordings. Features like eye-related events, mouth activity, head motion parameters, and camera-based HR estimation were investigated and classified into three classes: neutral, relaxed, and stressed/anxious. They got 91.68\% accuracy using the AdaBoost classifier. Kopaczka, Marcin, et al. \cite{kopaczka2019modular} developed a system consisting of face detection and facial landmark detection and employed Histogram of Oriented Gradients (HOG) features using a random forest classifier, achieving a classification accuracy of 65.75\% for four basic emotions (namely, Neutral-Happy-Sad-Surprised). \cite{zhang2020video} addressed the issue of stress detection using video-based camera sensors. They used facial expressions and action motions from videos to identify stress with a performance of 85.42\% accuracy for stress binary classification. Cardone, Daniela, et al. \cite{cardone2020driver} used thermal infrared images for stress detection. In this study, they achieved 80\% for the area under the curve (AUC), sensitivity of 77\%, and specificity of 78\% for binary classification based on a non-linear support vector regression (SVR). Jeon, Taejae, et al. \cite{jeon2020stress}  proposes a stress recognition algorithm using face images and facial landmarks to understand eye, mouth, and head movements. They achieved an accuracy of 64.63\%,  detecting three levels of no, weak, and strong stress.

\subsection{Multimodal Signals}
The SWELL project \cite{koldijk2016detecting} focuses on multimodal sensor data (computer logging, facial expressions, posture, and physiology). The study distinguishes neutral and stressful working conditions with 90\% accuracy using a support vector machine (SVM) model, using posture and facial expressions to provide valuable information. Bobade, Pramod, et al. \cite{bobade2020stress} used motion sensors alongside physiological signals (ECG, BVP, TEMP, RR, EMG, and EDA) to detect stress. The results showed that machine learning techniques achieve accuracies of up to 81.65\% and 93.20\% for emotion detection(amusement, baseline, and stress) and binary stress detection(stress vs. non-stress), respectively. Deep learning achieves accuracies of up to 84.32\% and 95.21\% for the two and three-class classifications. Seo, Wonju, et al. \cite{seo2022deep} used ECG, RR, and facial landmarks extracted from video data and achieved an accuracy of 73.3\% for binary classification and 54.4\% for three-level stress classification using multivariate deep neural networks. A multimodal AI-based framework is proposed by Walambe, Rahee, et al. \cite{walambe2021employing}. The methodology involves fusing data of facial expressions, posture, HR, and computer interaction (Mouse activity, Left click, etc.) to detect workload stress and achieve an accuracy of 96.09\%. Naegelin, Mara, et al. \cite{naegelin2023interpretable}  utilize mouse, keyboard, and HRV features to detect three levels of perceived stress, valence, and arousal, and they achieved a maximum F1 score of 77.5\%.

Some of these recent findings show the promise of multimodal approaches with respect to improving the accuracy and reliability of real-time stress detection. However, these studies are evaluated in small datasets and evaluated in laboratory conditions. More datasets and methods are needed to improve the efficacy of these methods. Table \ref{tab:1} summarizes some reviewed papers, listing the signals they used and their performance in terms of accuracy (Acc), area under the curve (AUC), precision (Prec), and F1 score.

\section{Methods}

\subsection{Dataset}
The EmpathicSchool \cite{hosseini2022empathicschool} dataset contains video-based facial expressions and corresponding physiological signals, HR, EDA, TEMP, and ACC from Empatica E4. 26 hours of data were collected from 20 participants from two universities, Tampere University, Finland, and the University of Louisiana at Lafayette, United States. The stress levels were determined via National Aeronautics and Space Administration Task Load Index questionnaires. The dataset includes seven different signal types, including both computer vision and physiological features. The video dataset provides facial landmarks. To the best of our knowledge, facial information of 68 landmarks has never been studied before. Additionally, 30 unique landmark features were incorporated to improve further stress prediction accuracy, such as the difference and distance of two or more landmark points on the face. 
 
\subsubsection{Biometric Data}

The following biometric signals were measured via the E4 wristband and are provided in the dataset:

• HR: Healthy resting heart rate (HR) ranges from 60 to 100 beats per minute, regardless of gender. However, HR is influenced by activity and emotional state. Wearable HR sensors enable the analysis of patterns and fluctuations, providing insights into stress, exertion, and cardiovascular health.

• TEMP: Skin temperature is influenced by skin blood flow and can reveal stress, fever, and changes in blood flow. Wearable ST sensors can capture data, enabling analysis of temperature patterns and correlation with other physiological parameters.

• EDA: EDA (Electrodermal Activity), also known as GSR (Galvanic Skin Response), is a signal that quantifies sweat gland activity and offers insights into emotional and psychological arousal. It reflects reactions to stress, excitement, anxiety, and other emotional states. Wearable EDA sensors detect changes in skin conductance, providing valuable information about an individual's physiological and emotional responses.

• Accelerometers (ACC): ACC data, measured by sensors, records object acceleration. These sensors have diverse applications, such as human action recognition and step counting. Wearable ACCs track movement patterns, offering information about activity levels.

\subsubsection{Facial Landmarks}
Facial landmarks are key points on a face that are used to identify features on a face, such as corners of eyes, eyebrows, tip of the nose, and lips. Image processing systems typically use these features to detect facial orientation, expression, or general facial patterns. 2D pixel values on the other hand are raw signals stored as pixel values of the entire image in a 2D array. Compared to a 2D image, facial landmarks provide a compact representation of facial features, capturing essential and relevant features. Extracting 68 facial landmarks is consisted of two steps face detection and identifying smaller features of the face, like eyes. We used Dlib 68-point face detection \cite{king2009dlib} with OpenCV \cite{opencv_library} to extract facial landmarks. Figure 1 shows 68 specific points on the face, which we refer to as landmarks. Analysis of spatial and temporal changes in these landmarks enables machine learning models to detect and classify stress.

\begin{figure}[h!]
       
       \centering
       \includegraphics[width=0.5\textwidth]{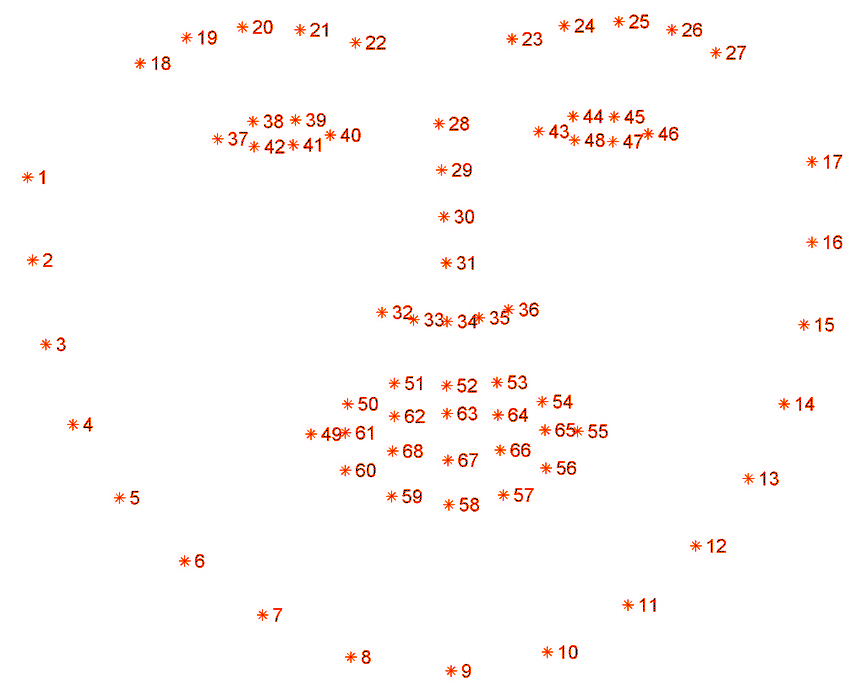}
       \caption{Illustration of the 68 facial landmarks}
       \label{fig:my_label}
   \end{figure}
\subsection{Pre-processing}
 The original signal data was collected with a frequency of 4 Hz in the EmphaticSchool \cite{hosseini2022empathicschool} dataset. We down-sample the data to a frequency of 1 Hz to reduce training and testing times. To reduce the effect of noise and artifacts on the stress data, we separated the EDA signal into its phasic and tonic components, and several features of each component were derived. Table \ref{tab:table2} shows the different features of EDA signals.
 
 \begin{table}[h!]
 
 \caption{EDA signals}
\centering
\begin{tabular}{|p{2cm}|p{7cm}|}
\hline
\textbf{EDA Signals}        & \multicolumn{1}{c|}{\textbf{Description}}             \\ \hline
EDA\_Clean        & Filtered version of the raw EDA signal        \\ \hline
EDA\_Tonic        & Sustained changes in skin conductance         \\ \hline
EDA\_Phasic       & Rapid changes in skin conductance             \\ \hline
SCR\_Onsets       & Starting points of SCR \\ \hline
SCR\_Peaks        & Highest points of SCR  \\ \hline
SCR\_Height       & Intensity or strength of the response     \\ \hline
SCR\_Amplitude    & Magnitude of the response                 \\ \hline
SCR\_RiseTime     & Duration for a SCR  \\ \hline
SCR\_Recovery     & Period after a SCR  \\ \hline
SCR\_Rec.Time     & Recovery time after an arousing event     \\ \hline
\end{tabular}
\label{tab:table2}
\end{table}

\begin{figure}[h!]
       
       \centering
       \includegraphics[width=0.8\textwidth]{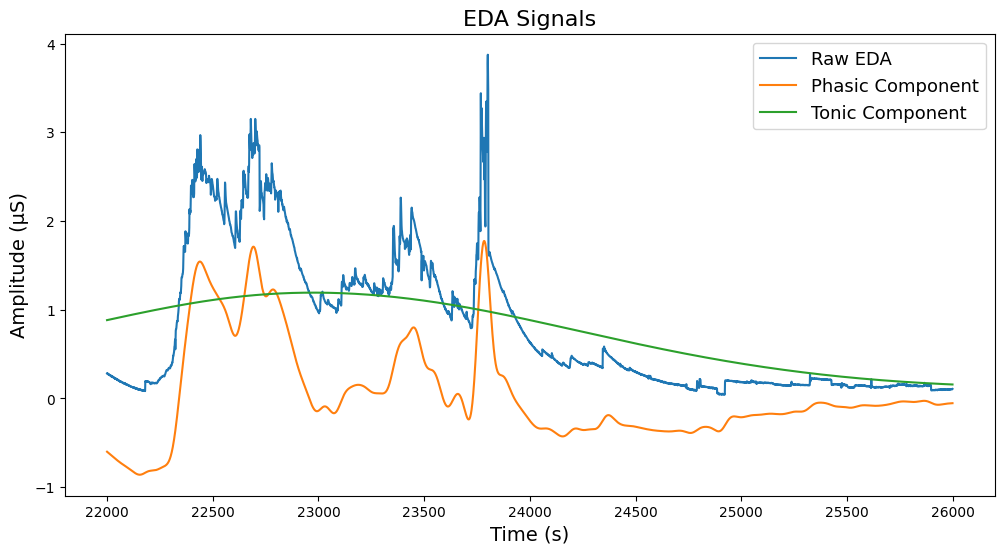}
       \caption{Phasic and Tonic components of raw EDA signal}
       \label{fig:fig2}
   \end{figure}

 The phasic component of the EDA signal represents the rapid and temporary changes in skin conductance. Phasic changes are characterized by short-duration and high-amplitude fluctuations in the EDA signal. The tonic component of the EDA signal represents the relatively slower and more sustained changes in skin conductance. Unlike phasic changes, the tonic component is not directly linked to specific events or stimuli but rather reflects the overall arousal level of an individual over a longer period. Figure \ref{fig:fig2} shows a part of the raw EDA signal with relative phasic and tonic components \cite{bailey2017electrodermal}.

 The statistical features of each signal are generated using a rolling window of 40 seconds with a sliding of 20 seconds due to the data characteristic, computational efficiency, and training data availability.
 The average stress level of each sliding window was obtained from the raw signal and binned into three different bins, namely no-stress 0(0 to 6), medium-stress 1(7 to 13), and stressful 2(14 to 19). The statistical features of each participant were extracted using the tsfresh \cite{christ2018time} Python library and normalized independently. The statistical features were extracted, and the missing data were removed. We used the data of 10 participants at different sessions for 26 hours. The dlib facial expressions model \cite{king2009dlib} consumes the video signals and generates the facial landmarks. We also extract several facial features and investigate the correlation of mentioned features using different feature selection methods. To evaluate their performance, the resulting multimodal data of biometric signals and facial landmarks consisted of 1897 features and the raw signals consumed by three deep learning networks: multivariate, early-fusion, and late-fusion. 
 
\subsubsection{Multivariate models}
 The signals and generated features were given to a deep-learning model. One hundred seventy-five features, including 15 statistical features (Absolute Energy, Fourier Entropy, Skewness, Autocorrelation, Quantile, kurtosis, Count Above Mean, Count Below Mean, Variation Coefficient, Root Mean Square, Variance, Mean, Standard Deviation, Maximum, Minimum ) of four different signals (EDA, Temp, HR, ACC), and 1904 features of 68 facial landmarks were generated and used in the feature selection model. Biometric signal and facial landmark features were decreased to 30 and 100 using lasso regularization \cite{tibshirani1996regression}, respectively.
 
\subsubsection{Multimodal Early-Fusion model}
The multimodal data comprising 2079 features combining biometric and facial landmarks data was decreased to 100 features using lasso regularization \cite{tibshirani1996regression}. The added unique features were found to be crucial in this study.

\subsubsection{Multimodal Late-Fusion model} 
The multimodal fusion model consumed the same data as our early-fusion networks, including 30 biometric and 100 landmark features with 1Hz frequency. The model showed promising results and was comparable to the early-fusion approach. In addition, the concatenation outperforms the early-fusion models.

\subsubsection{Feature analysis}
\begin{figure}
    \centering
    \includegraphics[width=.6\linewidth]{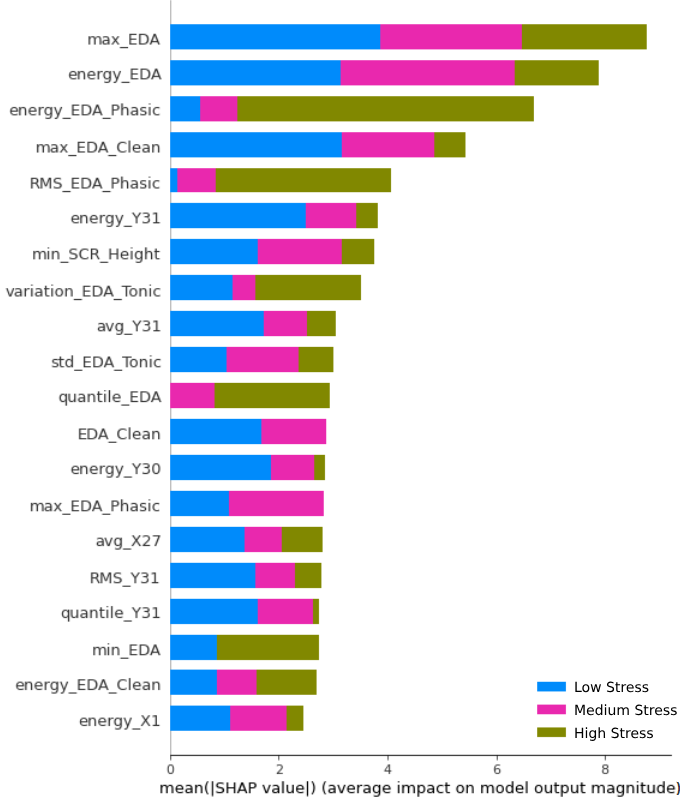}
    \caption{Explainable AI features and feature contribution for stress level discrimination}
    \label{fig:shap}
\end{figure}

\begin{figure*}[btp]
    \centering
    \includegraphics[width = \textwidth]{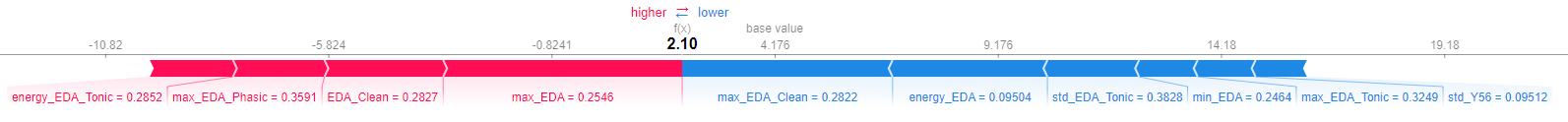}
    \caption{force plot SHAP values}
    \label{fig:force}
\end{figure*}

We employed three different types of feature selection, namely filter methods, namely Spearman \cite{myers2004spearman}, Pearson \cite{cohen2009pearson}, and Variance Threshold \cite{guyon2003introduction}, wrapper method namely Recursive Feature Elimination (RFE) \cite{guyon2002gene} and embedded methods of Random Forest (RF) Importance \cite{breiman2001random}, and Lasso Regularization \cite{tibshirani1996regression}. The top 10 features are presented in Table \ref{table3:tab3}. The Spearman and Pearson analyses yielded similar results, emphasizing facial landmark features. Conversely, the RF Importance and RFE analysis identified a combination of facial landmarks, ST, and EDA (Tonic and Phasic) features are the top contributors. The tonic component of EDA appears to have the highest variance using the variance threshold. Comparing the six feature selection methods, we observed that lasso regularization provided the most favorable outcome for running the models in this study, selecting features related to EDA and facial landmarks. The second and third best outcomes were from RF importance and RFE, respectively.
 We provided SHAP values and feature contributions using the Lasso regularization model. Figures \ref{fig:shap},\ref{fig:force} show the explainable information of the model using SHAP library \cite{lundberg2020local}. Among all features, the Max\_EDA had the maximum effect on the prediction results of all three stress levels, while some features, e.g., min\_EDA and quantile\_EDA, only discriminate between two stress levels. In Figure \ref{fig:force}, red signals represent the signals pushing higher stress levels positively associated with stress (e.g., Max\_EDA, EDA\_Clean, and phasic Max\_EDA). In contrast, the blue features represent lower stress levels negatively associated with stress (e.g., Max\_EDA\_clean, Energy\_EDA, standard deviation of tonic EDA). 

\begin{table}[h!]
\caption{Top ten features using different feature selection techniques}
\resizebox{\linewidth}{!}{
\begin{tabular}{|c|c|c|c|}

\hline
\textbf{Rank} & \textbf{Spearmans}                & \textbf{Pearson} & \textbf{Variance Threshold}   \\ \hline
1             & energy\_X2                        & energy\_X2       & below\_mean\_EDA\_Tonic       \\ \hline
2             & energy\_X1                        & energy\_X3       & above\_mean\_EDA\_Tonic       \\ \hline
3             & energy\_X3                        & energy\_X4       & skew\_EDA\_Tonic              \\ \hline
4             & energy\_X4                        & energy\_X1       & kurtosis\_EDA\_Tonic          \\ \hline
5             & RMS\_X1                           & RMS\_X2          & energy\_EDA\_Tonic            \\ \hline
6             & energy\_Y9                        & RMS\_X3          & max\_EDA\_Tonic               \\ \hline
7             & RMS\_X2                           & avg\_X2          & EDA\_Tonic                    \\ \hline
8             & avg\_X1                           & avg\_X3          & avg\_EDA\_Tonic               \\ \hline
9             & Y9                                & RMS\_X4          & RMS\_EDA\_Tonic               \\ \hline
10            & energy\_X5                        & RMS\_X1          & quantile\_EDA\_Tonic          \\ \hline
\textbf{Rank} & \textbf{RF Importance} & \textbf{RFE}     & \textbf{Lasso Regularization} \\ \hline
1             & energy\_X2                        & std\_EDA\_Clean                & max\_EDA                      \\ \hline
2             & energy\_EDA\_Tonic                & max\_EDA
                & max\_EDA\_Clean               \\ \hline
3             & max\_TEMP                         & energy\_EDA                & variance\_X21                 \\ \hline
4             & variation\_EDA\_Phasic            & variation\_EDA
                & std\_Y56                      \\ \hline
5             & skew\_EDA\_Tonic                  & energy\_X5                & energy\_Y31                   \\ \hline
6             & energy\_X32                       & energy\_X4                & energy\_EDA                   \\ \hline
7             & max\_EDA\_Clean                   & avg\_X27                & std\_X3                       \\ \hline
8             & energy\_EDA\_Phasic               & avg\_X25                & variation\_Y22                \\ \hline
9             & max\_Y24                          & avg\_X61                & variation\_Y28                \\ \hline
10            & RMS\_EDA\_Phasic                  & energy\_X65               & energy\_Y30                   \\ \hline
\end{tabular}
}
\label{table3:tab3}
\end{table}

\section{Stress detection models}
Using a sliding window to combine statistical features of data with a random train-test split can easily result in an overfit model. For stress detection with biometric signals, Leave One Out Cross Validation (LOOCV) is crucial. It prevents overfitting and offers a realistic gauge of the model's ability to generalize, as it tests the model with new data at each iteration. We fine-tuned our proposed stress detection network using the LOOCV approach to circumvent potential pitfalls. Moreover, stress detection models often encounter data from new individuals they were not trained on. LOOCV ensures the model's performance is evaluated reliably and meaningfully. Our investigation focused on three distinct network architectures: multivariate, early-fusion, and late-fusion.

\subsection{Multivariate}
We analyzed biometric signals and facial landmarks for feature extraction.  Then, each model consumed the features separately. The one-dimensional Convolutional Neural Network (1D-CNN), two-dimensional convolutional neural network (2D-CNN), and pure fully connected, deep neural network (FCDNN) was trained and tested using the LOOCV paradigm. By employing this approach, we aimed to obtain separate results for biometric signals and facial landmarks, allowing us to independently evaluate the models' performance for each type of data.

In our study, when using the 1D-CNN model to analyze facial landmarks, We treat the 68 facial landmarks as time series data, similar to how we handle ACC data that tracks coordinate changes over time. We focused on understanding how these landmarks change sequentially rather than capturing complex spatial relationships. On the other hand, using the 2D-CNN model, We pay more attention to how facial landmarks are arranged in space, which helps us capture overall spatial features.

\subsection{Early-Fusion}

\begin{figure}[h!]
       \centering
       \includegraphics[width=0.9\textwidth]{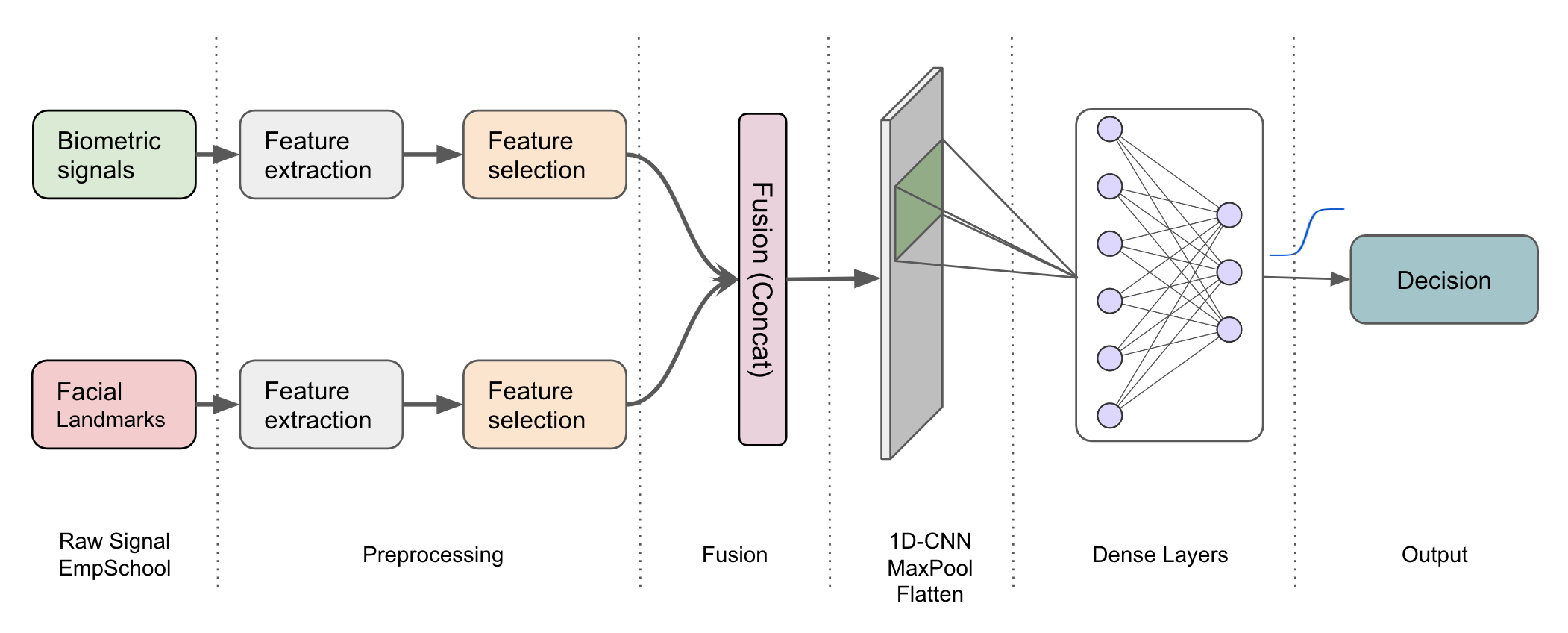}
       \caption{1D-CNN Early-Fusion Network }
       \label{fig:network1}
   \end{figure}

Figure \ref{fig:network1} shows the early-fusion stress detection model. The model consumes the concatenation of selected facial landmarks and biometric signal features to detect stress. We compared three early-fusion models, namely 1DCNN, 2D-CNN, and FCDNN. The 1DCNN and 2DCNN models use a one or two-dimensional convolutional layer to extract the spatiotemporal information of the representation of the signals for fully connected neural network layers, respectively. However, the FCDNN model directly consumes the concatenated features to detect stress. This integration enables the model to simultaneously analyze the biometric signals and visual information following the LOOCV paradigm.
  
\subsection{Late-Fusion}

\begin{figure}[h!]
       \centering
       \includegraphics[width=0.9\textwidth]{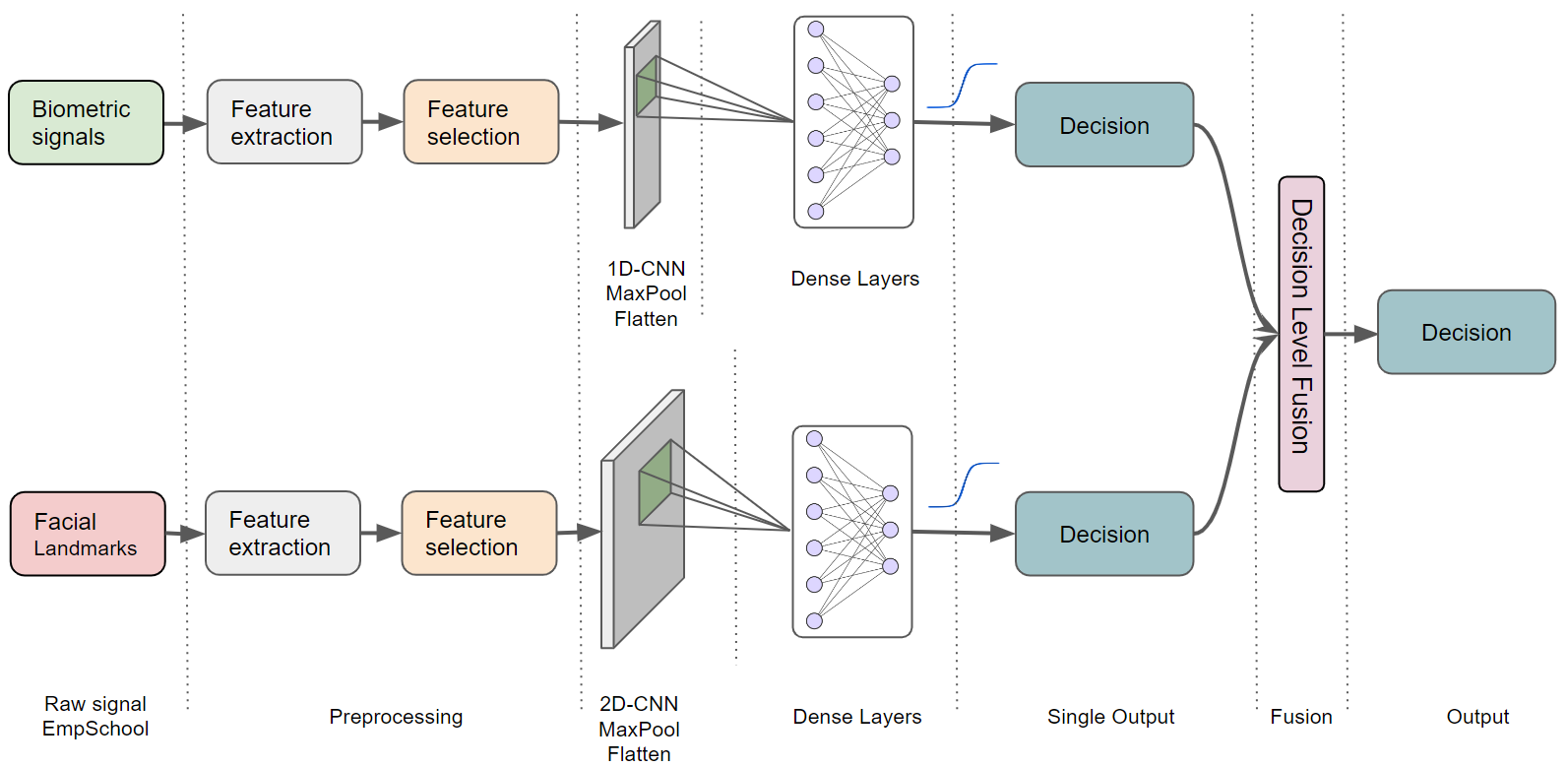}
       \caption{decision-level Fusion Network}
       \label{fig:network2}
   \end{figure}

\begin{figure}[h!]
       \centering
       \includegraphics[width=0.9\textwidth]{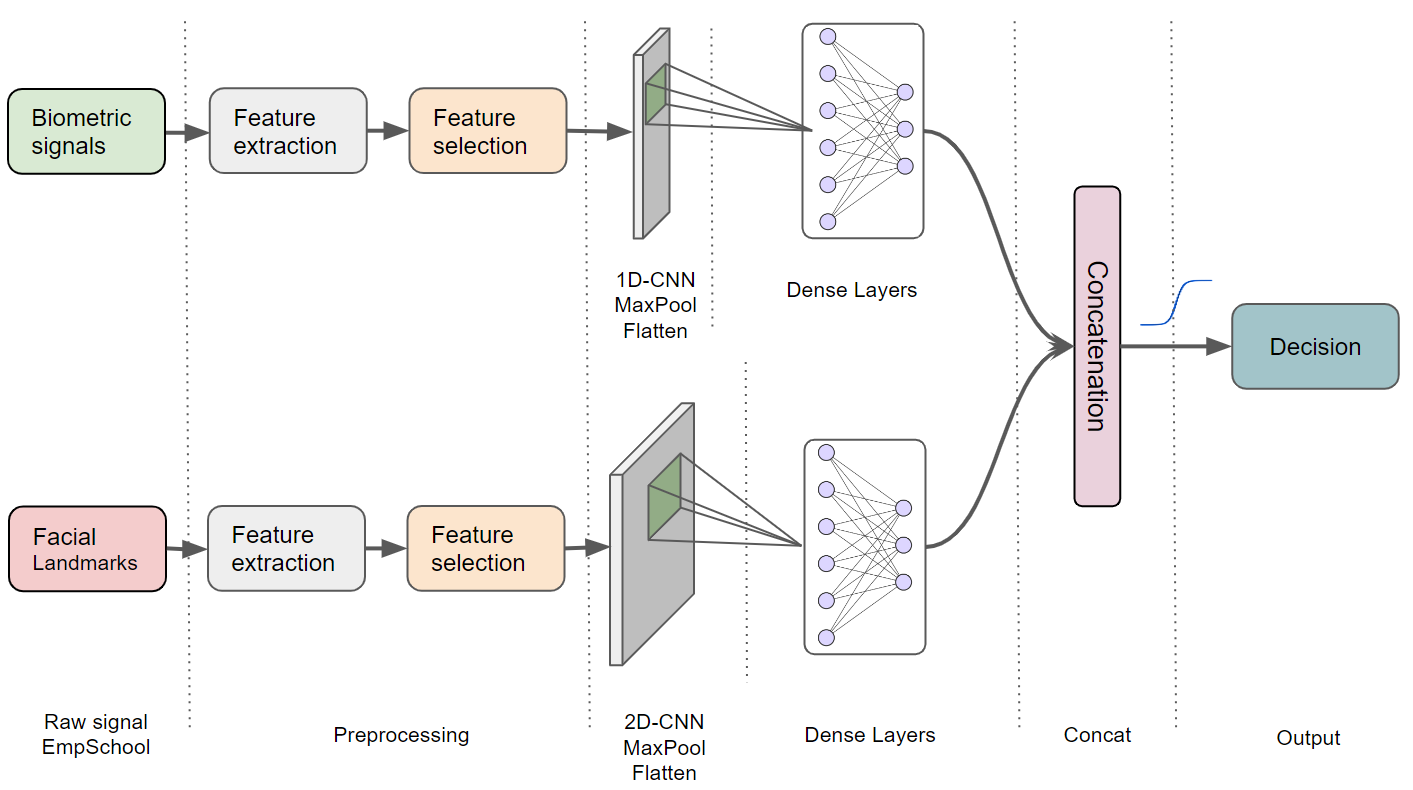}
       \caption{late-fusion Concatenation Network}
       \label{fig:network3}
   \end{figure}

The proposed network architecture Figure \ref{fig:network2} consists of two components that incorporate biometric signals and 68 landmarks in parallel. The late-fusion models consume the same features extracted using feature extraction modules. The biometric signals are fed into a 1D-CNN model, allowing us to capture temporal patterns and dependencies. Meanwhile, the landmarks data is processed using a 2D-CNN model to extract spatial information. We employ the decision-level Fusion technique to merge the results and produce a final output using the Softmax classifier (Figure \ref{fig:network2}). In the second experiment, we implemented a concatenation-based approach, wherein we fused the outputs of the two mentioned models, effectively giving rise to a novel composite model.
We also used a late-fusion concatenation model, combining one and two-dimension convolutional networks (Figure \ref{fig:network3}) before the decision layer (SoftMax). The model concatenates the representation of facial landmarks and biometric signals before the decision layer for stress detection. The concatenation late-fusion model showed promising results compared to other networks. However, the decision-level concatenation model outperforms the late-fusion network.

 We used accuracy, precision, recall, F1 score, and computational cost to perform model comparisons.

\begin{itemize}
    \item Accuracy: The ratio of the number of correct predictions to the total number of predictions 
        \begin{equation}
        Accuracy = \frac{TP+TN}{TP+TN+FP+FN} 
        \end{equation}

    \item Precision: Measures the proportion of correctly predicted positive instances out of all instances predicted as positive
        \begin{equation}
        Precision = \frac{TP}{TP+FP} 
        \end{equation}
Precision is a measure of how accurately the model can classify a specific class and in cases where the work is severely stressful, like military, firefighting, and nursing. 

    \item Recall: Measures the proportion of correctly predicted positive instances out of all actual positive instances
        \begin{equation}
        Recall = \frac{TP}{TP+FN} 
        \end{equation}
Recall is important when misclassifying the stress is not crucial (e.g., the wellness, meditation, and relaxation tasks). We prefer models with higher recall scores in stress treatment tasks.
        
    \item F1\_score: Provides a balanced measure of a classification model's performance by combining the precision and recall scores.
        \begin{equation}
        F1\_Score = 2\times\frac{Precision \times 
        Recall}{Precision+Recall}
        \end{equation}
\end{itemize}

TP, TN, FP, and FN are True Positive, True Negative, False Positive,  and False Negative, respectively.

F1\_score is needed in mental health applications where both stress onset and relaxation are important.  Furthermore, the models with higher f1\_score are desired in physiological signal analysis than the precision and recall score. 

\begin{table*}[]
\caption{Multimodal and multivariate neural network models performance (\%) comparison }
\begin{tabular}{|l|llllll|lll|ll|}
\hline
Network & \multicolumn{6}{c|}{Multi-Variate}                                                                                                                                 & \multicolumn{3}{c|}{Early-Fusion}                                                                                    & \multicolumn{2}{c|}{Late-Fusion}                                         \\ \hline
Model   & \multicolumn{2}{c|}{1D-CNN}                                 & \multicolumn{2}{c|}{2D-CNN}                                 & \multicolumn{2}{c|}{FCDNN}             & \multicolumn{1}{l|}{\multirow{2}{*}{1D-CNN}} & \multicolumn{1}{l|}{\multirow{2}{*}{2D-CNN}} & \multirow{2}{*}{FCDNN} & \multicolumn{1}{l|}{\multirow{2}{*}{Decision}} & \multirow{2}{*}{Concat} \\ \cline{1-7}
Data    & \multicolumn{1}{l|}{BIO}     & \multicolumn{1}{l|}{LND}     & \multicolumn{1}{l|}{BIO}     & \multicolumn{1}{l|}{LND}     & \multicolumn{1}{l|}{BIO}     & LND     & \multicolumn{1}{l|}{}                        & \multicolumn{1}{l|}{}                        &                        & \multicolumn{1}{l|}{}                          &                         \\ \hline
Acc     & \multicolumn{1}{l|}{93.01} & \multicolumn{1}{l|}{91.09} & \multicolumn{1}{l|}{91.28} & \multicolumn{1}{l|}{95.02} & \multicolumn{1}{l|}{92.08} & 92.42 & \multicolumn{1}{l|}{98.38}                 & \multicolumn{1}{l|}{94.13}                 & 94.62                & \multicolumn{1}{l|}{94.39}                   & 92.67                 \\ \hline
Prc     & \multicolumn{1}{l|}{90.00} & \multicolumn{1}{l|}{86.30} & \multicolumn{1}{l|}{88.30} & \multicolumn{1}{l|}{88.30} & \multicolumn{1}{l|}{87.33} & 83.51 & \multicolumn{1}{l|}{96.73}                 & \multicolumn{1}{l|}{93.65}                 & 93.08                & \multicolumn{1}{l|}{94.94}                   & 91.11                 \\ \hline
Rec     & \multicolumn{1}{l|}{85.44} & \multicolumn{1}{l|}{83.80} & \multicolumn{1}{l|}{84.91} & \multicolumn{1}{l|}{86.74} & \multicolumn{1}{l|}{84.80} & 81.52 & \multicolumn{1}{l|}{95.61}                 & \multicolumn{1}{l|}{87.31}                 & 89.26                & \multicolumn{1}{l|}{86.35}                   & 86.53                 \\ \hline
F1      & \multicolumn{1}{l|}{85.71} & \multicolumn{1}{l|}{82.62} & \multicolumn{1}{l|}{84.83} & \multicolumn{1}{l|}{86.45} & \multicolumn{1}{l|}{84.92} & 81.91 & \multicolumn{1}{l|}{95.70}                 & \multicolumn{1}{l|}{88.63}                 & 90.12                & \multicolumn{1}{l|}{88.63}                   & 87.54                 \\ \hline
\end{tabular}
\label{tab:table4}
\end{table*}

\section{Results}
The performance of trained deep neural networks with regard to multi-variate, early, and late-fusion approaches was evaluated on facial landmarks, biometric signals, and a combination of their features. We compared the performance and computational costs of the networks using training and testing times. We trained and tested ten networks using six feature selection methods and compared the results of the networks in detail. Furthermore, Table \ref{tab:table4} summarizes the performance of each network in terms of the above-mentioned metrics.

\subsection{Multivariate Results}
In the first experiment, we use the standard multivariate signals for a single neural network architecture. Biometric signals are multi-variate, but each signal sequence is still 1-dimensional, and 1DCNN has been shown to capture patterns within the individual signals and cross-channel relationships \cite{kiranyaz20191,khoshkhahtinat2023multi}. The 1D-CNN model captures temporal patterns and biometric signals, while the 2D-CNN extracts spatial features of facial landmarks. We can observe in Table 4 that 1D-CNN provides good performance for biometric signals. This is on par with prior studies where we achieved 93.01\% accuracy \cite{padmaja2018machine,shu2020wearable,hsu2020automatic,sharma2020automated,zhu2022feasibility,zhu2022multimodal,kuttala2023multimodal,schmidt2018introducing}. Landmark data (LND) on the other hand are vector data (like images), and we observe that 2D-CNN models provide better performance for facial landmarks. The 2D CNN results are close to what is observed in \cite{giannakakis2017stress,zhang2020video,cardone2020driver,kopaczka2019modular,jeon2020stress}. Fully connected DNN models, on the other hand, are good at capturing non-linear relationships, which has good overall performance when you look at both signals. Fully connected DNN models treat data in their true form and do not capture temporal dependencies, making these models vulnerable to capturing false positives as true. In other words, the fully connected DNN model cannot distinguish between decreasing and increasing trends. For example, when stress has passed, the signs from the tonic EDA signal still remain. As expected, the 1D-CNN model works better with biometric signals, resulting in 93.01\% accuracy, and 2D-CNN works best with facial landmarks, showing 95.02\% accuracy compared to fully connected deep neural network model with around 92\% for both biometric and facial landmarks signals.
\subsection{Late-Fusion Results}
In the case of late fusion, each modality is processed independently, and their outputs are combined at a later stage, usually closer to the final decision or prediction stage. The main advantage of this approach is its flexibility because each modality is processed independently, accommodating the unique characteristics of each data source. Also, the spatial and temporal scales need not perfectly align for late-fusion. We experiment with late-fusion models that combine one and two-dimensional convolutional neural networks. The first late-fusion approach involved combining the outcomes of the 1D-CNN and 2D-CNN models through decision-level fusion. We can observe in Table-4 that results from late-fusion are better than the multi-variate model but lower than early-fusion for both decision-based fusion and concatenation-based fusion. 

The concatenation-based approach concatenates the input and output of every mentioned model into a single composite model. The fusion of two distinct models, each specialized in a different modality, contributed to the overall accuracy of 94.39\% and 92.67\% for stress detection using decision-level and concatenation fusion, respectively. We can observe that the late fusion model provides better performance compared to the multivariate model, but the model performance is slightly lower compared to the 1D CNN-based early fusion model. 
\subsection{Early-Fusion Results}
\begin{figure}[]
       \centering
       \includegraphics[width=0.4\textwidth]{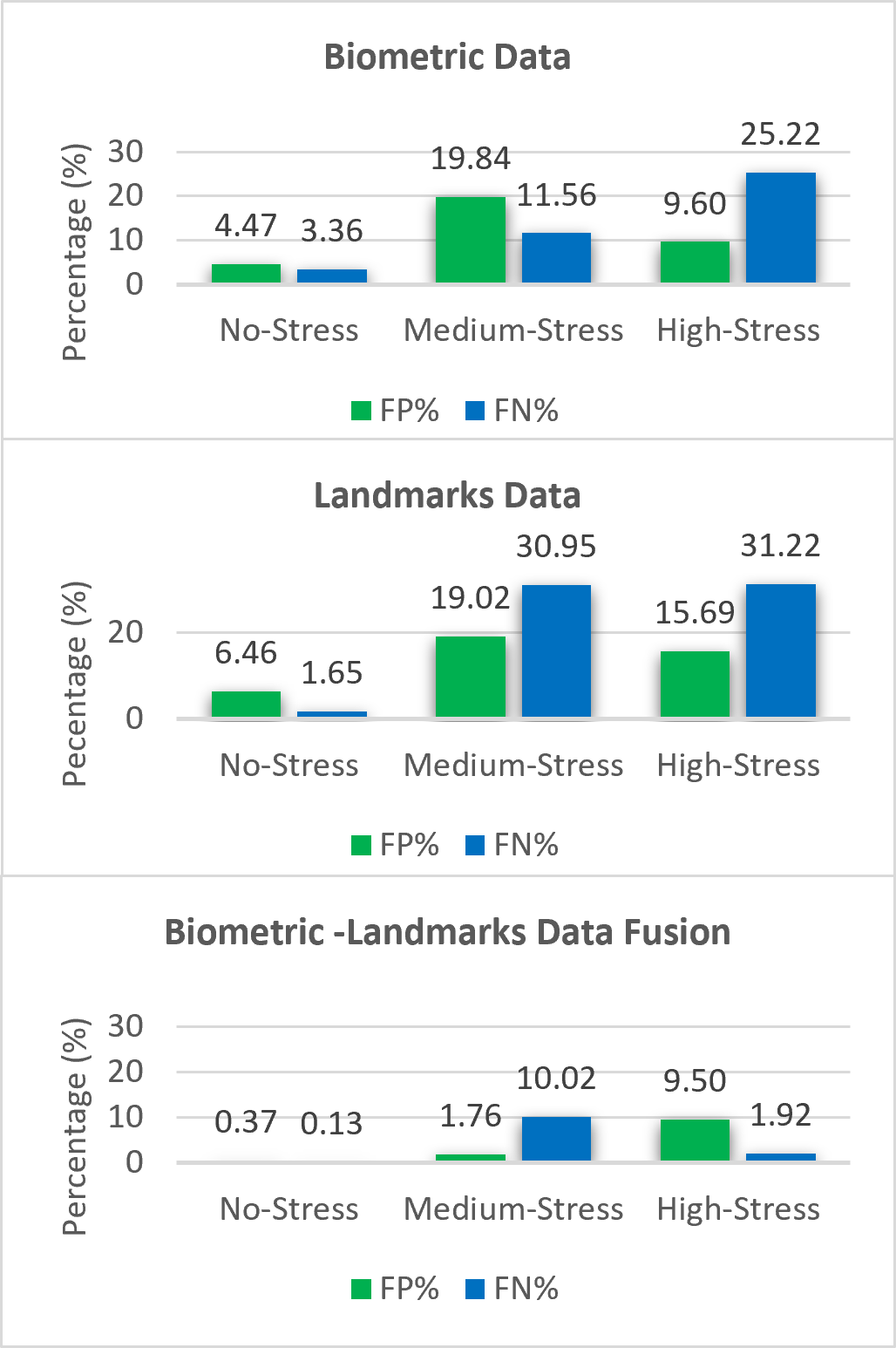}
       \caption{FP, and FN rate for each class in 1D-CNN Early-Fusion Network }
       \label{fig:fpfn}
   \end{figure}

Early fusion, or feature-level fusion, combines features at the early processing stage in the end-to-end model architecture. This is most suitable when data from modalities correspond to the same instance or context \cite{dong2014performance}. Stress signals may exhibit some form of temporal alignment, but not always. There is high inter-subject variability with respect to how individuals perceive and react to stress \cite{healey2005detecting}. Also, acute stress leads to immediate and observable physiological and behavioral changes \cite{morgado2018impact} compared to chronic stress resulting from subtle and long-term changes. It is also observed by \cite{soni2020review, hosseini2022development} that physiological signals such as HRV, skin conductance, and facial expressions due to the stressful event also occur within the close temporal context. In our case, the dataset includes two individual signals that have been shown to have temporal proximity. We observe that Early fusion performs better with accuracy, recall, and F1-score in most cases for 1D-CNN, 2D-CNN, and FCDNN compared to multi-variate models, but 1D-CNN-based early fusion has the best performance overall. The overall accuracy is 98.38\%, which I better compared to previous studies which are based on multi-variate signals \cite{seo2022deep,walambe2021employing, naegelin2023interpretable,hosseini2022empathicschool,koldijk2016detecting,bobade2020stress}. In the context of this dataset, the model is able to focus on the most relevant and informative features from both modalities. The analysis of False Positive (FP) and False Negative (FN) rates for biometric data, landmark data, and their combination (fusion) for the 1D-CNN model is illustrated in Figure\ref{fig:fpfn}. Using biometric signals and facial landmarks showed good accuracy in identifying non-stressful instances but did not perform well in distinguishing stressed instances. However, when both signals are combined, the model achieves better accuracy in identifying all three instances.
Reducing the FN rate is essential to ensure no high-stress situations are overlooked and individuals receive appropriate warning signals. The combined biometric and landmark data fusion approach showed promising results in decreasing the FN rate and enhancing the reliability of the stress detection system for individuals experiencing high stress.

We investigated various feature selection methods, and the following points summarize the results with the 1D-CNN model performance on combining both data types. Some of the results are as follows:
\begin{itemize}
\item Using RF importance resulted in an Accuracy of 96.94\%, Precision of 95.72\%, Recall of 91.20\%, and F1-score of 92.93\%. 
\item Using RFE resulted in an Accuracy of 96.63\%, Precision of 95.34\%, Recall of 92.54\%, and F1-score of 92.97\%. 
\item Using Variance Threshold resulted in an Accuracy of 92.76\%, Precision of 92.40\%, Recall of 87.52\%, and F1-score of 88.19\%. 
\end{itemize}
We investigated the importance of using the Tonic and Phasic components of the EDA signal. Our experiment showed that using these components in the data in the 1D-CNN model resulted in 3.57\%, 9.23\%, 10.47\%, and 10.71\% higher Accuracy, Precision, Recall, and F1-score, respectively.

\begin{table}[]
\caption{The computational cost of the models}
\begin{adjustbox}{width=\textwidth}
\begin{tabular}{|l|llllll|lll|ll|}
\hline
Network                & \multicolumn{6}{c|}{Multi-Variate}                                                                                                                                            & \multicolumn{3}{c|}{Early-Fusion}                                                                                                         & \multicolumn{2}{c|}{Late-Fusion}                                                              \\ \hline
Model                  & \multicolumn{2}{c|}{1D-CNN}                               & \multicolumn{2}{c|}{2D-CNN}                              & \multicolumn{2}{c|}{FCDNN}                             & \multicolumn{1}{c|}{\multirow{2}{*}{1D-CNN}} & \multicolumn{1}{c|}{\multirow{2}{*}{2D-CNN}} & \multicolumn{1}{c|}{\multirow{2}{*}{FCDNN}} & \multicolumn{1}{c|}{\multirow{2}{*}{decision}} & \multicolumn{1}{c|}{\multirow{2}{*}{Concat}} \\ \cline{1-7}
Data                   & \multicolumn{1}{c|}{BIO}    & \multicolumn{1}{c|}{LND}    & \multicolumn{1}{c|}{Bio}   & \multicolumn{1}{c|}{LND}    & \multicolumn{1}{c|}{BIO}    & \multicolumn{1}{c|}{LND} & \multicolumn{1}{c|}{}                        & \multicolumn{1}{c|}{}                        & \multicolumn{1}{c|}{}                       & \multicolumn{1}{c|}{}                          & \multicolumn{1}{c|}{}                        \\ \hline
Trainable   Parameters & \multicolumn{1}{l|}{23,931} & \multicolumn{1}{l|}{79,931} & \multicolumn{1}{l|}{4,923} & \multicolumn{1}{l|}{27,323} & \multicolumn{1}{l|}{32,703} & 46,703                   & \multicolumn{1}{l|}{79,931}                  & \multicolumn{1}{l|}{27,323}                  & 46,703                                      & \multicolumn{1}{l|}{51,254}                    & 51,176                                       \\ \hline
Train Time   (s)       & \multicolumn{1}{l|}{2,610}  & \multicolumn{1}{l|}{2,590}  & \multicolumn{1}{l|}{2,870} & \multicolumn{1}{l|}{2,930}  & \multicolumn{1}{l|}{2,290}  & 2,220                    & \multicolumn{1}{l|}{2,820}                   & \multicolumn{1}{l|}{2,870}                   & 2,310                                       & \multicolumn{1}{l|}{5,810}                     & 3,850                                        \\ \hline
Test Time   (s)        & \multicolumn{1}{l|}{4.19}   & \multicolumn{1}{l|}{4.01}   & \multicolumn{1}{l|}{3.4}   & \multicolumn{1}{l|}{3.65}   & \multicolumn{1}{l|}{3.27}   & 3.14                     & \multicolumn{1}{l|}{4.07}                    & \multicolumn{1}{l|}{3.64}                    & 3.35                                        & \multicolumn{1}{l|}{3.73}                      & 5.86                                         \\ \hline
\end{tabular}
\end{adjustbox}
\label{Table:table5}
\end{table}

\subsection{Computational Cost}
Table \ref{Table:table5} presents the computational cost of these models with respect to the trainable parameters, training time, and prediction time. We used a two-28-core (56 CPUs) node with Intel(R) Xeon(R) CPU E5-2660 v4 @ 2.00GHz CPU and L1d, L1i, L2, and L3 cache of 32K, 32K, 256K, 35840K, respectively. The node used a P100 Tesla GPU for deep-learning computations. Early-fusion models show lower training and test times, while late-fusion models require more training time. The early fusion model needs lower training time due to the model only needs to be trained once on the combined feature set. However, in the late-fusion model, Each model needs to be trained separately, which can take more time, especially because there are many features. Moreover,  The complexity of the overall system is higher because it includes two different models. The 2D-CNN model on biometric signals had the lowest number of parameters. Because all the signals were concatenated before the convolutional model, it resulted in lower training time and accuracies due to not capturing enough information. The FDCNN model using facial landmarks requires lesser training time than late fusion models; however, the overall performance of these models drops about 8\%. However, the 1D-CNN model has the maximum number of trainable parameters but has a lower training time than the late-fusion level neural networks. The 1D-CNN model is the most accurate with combined data but has the most trainable parameters. The balance between accuracy, complexity, and computational efficiency varies across models, showing the importance of modality fusion and pre-processing choices for effective stress detection.
\section{Conclusion}
This paper studies stress detection through multimodal learning that encompasses biometric signals and facial landmarks. We used the EmpathicSchool dataset that contains video-based facial expressions and physiological signals when subjects were asked to perform stressful student tasks. We experimented with various early and late-fusion techniques based on 1-D CNN, 2D-CNN, and fully connected DNN models for multi-modal stress detection. We observe that early-fusion techniques with a 1D-CNN model provide the best performance with an accuracy of 98.38\%. This model performs better than multi-variate unimodal or multi-modal signals with late fusion. The fully connected DNN demonstrates strong performance, exceeding the 1D-CNN model on landmarks data alone and enhancing overall precision. Conversely, the 2D-CNN model performed well with facial landmarks data due to its ability to capture both spatial relationships and temporal dynamics of facial landmarks.

This study's results are essential in creating reliable stress detection methods, especially in healthcare, education, and the workplace. The study should be further expanded to include a more diverse population and tasks. The source code and datasets used in this study are for public use through the University of Louisiana CPHS-lab GitHub repository \cite{CPHS2020Majid}.


\begin{thebibliography}{9}
\bibitem[1]{cohen2007psychological}
Cohen, S., Janicki-Deverts, D., \& Miller, G. E. Psychological stress and disease. {\em Jama} {\bf 2007}, {\em 298}(14), 1685--1687.

\bibitem[2]{ghazal2021iot}
Ghazal, T. M., Hasan, M. K., Alshurideh, M. T., Alzoubi, H. M., Ahmad, M., Akbar, S. S., Al Kurdi, B., \& Akour, I. A. IoT for smart cities: Machine learning approaches in smart healthcare—A review. {\em Future Internet} {\bf 2021}, {\em 13}(8), 218.

\bibitem[3]{morgado2018impact}
Morgado, P., \& Cerqueira, J. J. The impact of stress on cognition and motivation. {\em Frontiers in Behavioral Neuroscience} {\bf 2018}, {\em 12}, 326.

\bibitem[4]{soni2020review}
Soni, A., \& Rawal, K. A review on physiological signals: Heart rate variability and skin conductance. In {\em Proceedings of First International Conference on Computing, Communications, and Cyber-Security (IC4S 2019)} {\bf 2020}, 387--399.

\bibitem[5]{hosseini2022multimodal}
Hosseini, S., Gottumukkala, R., Katragadda, S., Bhupatiraju, R. T., Ashkar, Z., Borst, C. W., \& Cochran, K. A multimodal sensor dataset for continuous stress detection of nurses in a hospital. {\em Scientific Data} {\bf 2022}, {\em 9}(1), 255.

\bibitem[6]{kiranyaz20191}
Kiranyaz, S., Ince, T., Abdeljaber, O., Avci, O., \& Gabbouj, M. 1-D convolutional neural networks for signal processing applications. In {\em ICASSP 2019-2019 IEEE International Conference on Acoustics, Speech and Signal Processing (ICASSP)} {\bf 2019}, 8360--8364.

\bibitem[7]{opencv_library}
Bradski, G. The OpenCV Library. {\em Dr. Dobb's Journal of Software Tools} {\bf 2000}.

\bibitem[8]{king2009dlib}
King, D. E. Dlib-ml: A Machine Learning Toolkit. {\em Journal of Machine Learning Research} {\bf 2009}, {\em 10}, 1755–1758.

\bibitem[9]{healey2005detecting}
Healey, J. A., \& Picard, R. W. Detecting stress during real-world driving tasks using physiological sensors. {\em IEEE Transactions on intelligent transportation systems} {\bf 2005}, {\em 6}(2), 156--166.

\bibitem[10]{dong2014performance}
Dong, Y., Gao, S., Tao, K., Liu, J., \& Wang, H. Performance evaluation of early and late fusion methods for generic semantics indexing. {\em Pattern Analysis and Applications} {\bf 2014}, {\em 17}, 37--50.

\bibitem[11]{bobade2020stress}
Bobade, P., \& Vani, M. Stress detection with machine learning and deep learning using multimodal physiological data. In {\em 2020 Second International Conference on Inventive Research in Computing Applications (ICIRCA)} {\bf 2020}, 51--57.

\bibitem[12]{zhang2022real}
Zhang, J., Yin, H., Zhang, J., Yang, G., Qin, J., \& He, L. Real-time mental stress detection using multimodality expressions with a deep learning framework. {\em Frontiers in Neuroscience} {\bf 2022}, {\em 16}, 947168.

\bibitem[13]{islam2023personalized}
Islam, T., \& Washington, P. Personalized Prediction of Recurrent Stress Events Using Self-Supervised Learning on Multimodal Time-Series Data. {\em arXiv preprint arXiv:2307.03337} {\bf 2023}.

\bibitem[14]{hosseini2022development}
Hosseini, L., Sharif Nia, H., \& Ashghali Farahani, M. Development and psychometric evaluation of family caregivers’ hardiness scale: a sequential-exploratory mixed-method study. {\em Frontiers in Psychology} {\bf 2022}, {\em 13}, 807049.

\bibitem[15]{hosseini2022empathicschool}
Hosseini, M., Sohrab, F., Gottumukkala, R., Bhupatiraju, R. T., Katragadda, S., Raitoharju, J., Iosifidis, A., \& Gabbouj, M. EmpathicSchool: A multimodal dataset for real-time facial expressions and physiological data analysis under different stress conditions. {\em arXiv preprint arXiv:2209.13542} {\bf 2022}.

\bibitem[16]{montesinos2019multi}
Montesinos, V., Dell’Agnola, F., Arza, A., Aminifar, A., \& Atienza, D. Multi-modal acute stress recognition using off-the-shelf wearable devices. In {\em 2019 41st Annual International Conference of the IEEE Engineering in Medicine and Biology Society (EMBC)} {\bf 2019}, 2196--2201.

\bibitem[17]{kuttala2023multimodal}
Kuttala, R., Subramanian, R., \& Oruganti, V. R. M. Multimodal Hierarchical CNN Feature Fusion for Stress Detection. {\em IEEE Access} {\bf 2023}, {\em 11}, 6867--6878.

\bibitem[18]{CPHS2020Majid}
Majid Satya Ravi. Stress-Detection-in-Nurse. {\em GitHub} {\bf 2021}.

\bibitem[19]{sioni2015stress}
Sioni, R., \& Chittaro, L. Stress detection using physiological sensors. {\em Computer} {\bf 2015}, {\em 48}(10), 26--33.


\bibitem[20]{christ2018time}
Christ, M., Braun, N., Neuffer, J., \& Kempa-Liehr, A. W. Time series feature extraction on basis of scalable hypothesis tests (tsfresh--a python package). {\em Neurocomputing} {\bf 2018}, {\em 307}, 72--77.

\bibitem[21]{ho1995random}
Ho, T. K. Random decision forests. In {\em Proceedings of 3rd international conference on document analysis and recognition} {\bf 1995}, 278--282.

\bibitem[22]{kleinbaum2002logistic}
Kleinbaum, D. G., Dietz, K., Gail, M., Klein, M., \& Klein, M. Logistic regression. {\em Springer} {\bf 2002}.

\bibitem[23]{chen2015xgboost}
Chen, T., He, T., Benesty, M., Khotilovich, V., Tang, Y., Cho, H., Chen, K., Mitchell, R., Cano, I., Zhou, T., et al. Xgboost: extreme gradient boosting. {\em R package version 0.4-2} {\bf 2015}, {\em 1}(4), 1--4.

\bibitem[24]{noble2006support}
Noble, W. S. What is a support vector machine? {\em Nature biotechnology} {\bf 2006}, {\em 24}(12), 1565--1567.

\bibitem[25]{windeatt2006accuracy}
Windeatt, T. Accuracy/diversity and ensemble MLP classifier design. {\em IEEE Transactions on Neural Networks} {\bf 2006}, {\em 17}(5), 1194--1211.

\bibitem[26]{rosenblatt1958perceptron}
Rosenblatt, F. The perceptron: a probabilistic model for information storage and organization in the brain. {\em Psychological review} {\bf 1958}, {\em 65}(6), 386.

\bibitem[27]{montavon2018methods}
Montavon, G., Samek, W., \& M{\"u}ller, K. R. Methods for interpreting and understanding deep neural networks. {\em Digital signal processing} {\bf 2018}, {\em 73}, 1--15.

\bibitem[28]{alberdi2016towards}
Alberdi, A., Aztiria, A., \& Basarab, A. Towards an automatic early stress recognition system for office environments based on multimodal measurements: A review. {\em Journal of biomedical informatics} {\bf 2016}, {\em 59}, 49--75.

\bibitem[29]{fukazawa2020smartphone}
Fukazawa, Y., Yamamoto, N., Hamatani, T., Ochiai, K., Uchiyama, A., \& Ohta, K. Smartphone-based mental state estimation: A survey from a machine learning perspective. {\em Journal of Information Processing} {\bf 2020}, {\em 28}, 16--30.

\bibitem[30]{zhai2006stress}
Zhai, J., \& Barreto, A. Stress detection in computer users based on digital signal processing of noninvasive physiological variables. In {\em 2006 international conference of the IEEE engineering in medicine and biology society} {\bf 2006}, 1355--1358.

\bibitem[31]{kurniawan2013stress}
Kurniawan, H., Maslov, A. V., \& Pechenizkiy, M. Stress detection from speech and galvanic skin response signals. In {\em Proceedings of the 26th IEEE International Symposium on Computer-Based Medical Systems} {\bf 2013}, 209--214.

\bibitem[32]{abouelenien2016human}
Abouelenien, M., Burzo, M., \& Mihalcea, R. Human acute stress detection via integration of physiological signals and thermal imaging. In {\em Proceedings of the 9th ACM international conference on pervasive technologies related to assistive environments} {\bf 2016}, 1--8.



\bibitem[33]{wijsman2011towards}
Wijsman, J., Grundlehner, B., Liu, H., Hermens, H., \& Penders, J. Towards mental stress detection using wearable physiological sensors. In {\em 2011 Annual International Conference of the IEEE Engineering in Medicine and Biology Society} {\bf 2011}, 1798--1801.

\bibitem[34]{sandulescu2015stress}
Sandulescu, V., Andrews, S., Ellis, D., Bellotto, N., \& Mozos, O. M. Stress detection using wearable physiological sensors. In {\em Artificial Computation in Biology and Medicine: International Work-Conference on the Interplay Between Natural and Artificial Computation, IWINAC 2015, Elche, Spain, June 1-5, 2015, Proceedings, Part I 6} {\bf 2015}, 526--532.

\bibitem[35]{schmidt2018introducing}
Schmidt, P., Reiss, A., Duerichen, R., Marberger, C., \& Van Laerhoven, K. Introducing wesad, a multimodal dataset for wearable stress and affect detection. In {\em Proceedings of the 20th ACM international conference on multimodal interaction} {\bf 2018}, 400--408.

\bibitem[36]{koldijk2016detecting}
Koldijk, S., Neerincx, M. A., \& Kraaij, W. Detecting work stress in offices by combining unobtrusive sensors. {\em IEEE Transactions on affective computing} {\bf 2016}, {\em 9}(2), 227--239.

\bibitem[37]{rizwan2019design}
Rizwan, M. F., Farhad, R., Mashuk, F., Islam, F., \& Imam, M. H. Design of a biosignal based stress detection system using machine learning techniques. In {\em 2019 international conference on robotics, electrical and signal processing techniques (ICREST)} {\bf 2019}, 364--368.

\bibitem[38]{giannakakis2017stress}
Giannakakis, G., Pediaditis, M., Manousos, D., Kazantzaki, E., Chiarugi, F., Simos, P. G., Marias, K., \& Tsiknakis, M. Stress and anxiety detection using facial cues from videos. {\em Biomedical Signal Processing and Control} {\bf 2017}, {\em 31}, 89--101.

\bibitem[39]{garcia2015automatic}
Garcia-Ceja, E., Osmani, V., \& Mayora, O. Automatic stress detection in working environments from smartphones’ accelerometer data: a first step. {\em IEEE journal of biomedical and health informatics} {\bf 2015}, {\em 20}(4), 1053--1060.

\bibitem[40]{gjoreski2016continuous}
Gjoreski, M., Gjoreski, H., Lu{\v{s}}trek, M., \& Gams, M. Continuous stress detection using a wrist device: in laboratory and real life. In {\em proceedings of the 2016 ACM international joint conference on pervasive and ubiquitous computing: Adjunct} {\bf 2016}, 1185--1193.

\bibitem[41]{kim2008emotion}
Kim, J., \& Andr{\'e}, E. Emotion recognition based on physiological changes in music listening. {\em IEEE transactions on pattern analysis and machine intelligence} {\bf 2008}, {\em 30}(12), 2067--2083.

\bibitem[42]{padmaja2018machine}
Padmaja, B., Prasad, V. V. R., \& Sunitha, K. V. N. A machine learning approach for stress detection using a wireless physical activity tracker. {\em International Journal of Machine Learning and Computing} {\bf 2018}, {\em 8}(1), 33--38.

\bibitem[43]{chen2021emotion}
Chen, S., Jiang, K., Hu, H., Kuang, H., Yang, J., Luo, J., Chen, X., \& Li, Y. Emotion recognition based on skin potential signals with a portable wireless device. {\em Sensors} {\bf 2021}, {\em 21}(3), 1018.

\bibitem[44]{feng2020investigating}
Feng, X., Lu, X., Li, Z., Zhang, M., Li, J., \& Zhang, D. Investigating the Physiological Correlates of Daily Well-being: A PERMA Model-Based Study. {\em The Open Psychology Journal} {\bf 2020}, {\em 13}(1).

\bibitem[45]{shu2020wearable}
Shu, L., Yu, Y., Chen, W., Hua, H., Li, Q., Jin, J., \& Xu, X. Wearable emotion recognition using heart rate data from a smart bracelet. {\em Sensors} {\bf 2020}, {\em 20}(3), 718.

\bibitem[46]{hsu2020automatic}
Hsu, Y. L., Wang, J. S., Chiang, W. C., \& Hung, C. H. Automatic ECG-Based Emotion Recognition in Music Listening. {\em IEEE Transactions on Affective Computing} {\bf 2020}, {\em 11}(1), 85--99.

\bibitem[47]{siirtola2019continuous}
Siirtola, P. Continuous stress detection using the sensors of commercial smartwatch. In {\em Adjunct Proceedings of the 2019 ACM International Joint Conference on Pervasive and Ubiquitous Computing and Proceedings of the 2019 ACM International Symposium on Wearable Computers} {\bf 2019}, 1198--1201.

\bibitem[48]{sharma2020automated}
Sharma, R., Pachori, R. B., \& Sircar, P. Automated emotion recognition based on higher order statistics and deep learning algorithm. {\em Biomedical Signal Processing and Control} {\bf 2020}, {\em 58}, 101867.

\bibitem[49]{zhu2022feasibility}
Zhu, L., Ng, P. C., Yu, Y., Wang, Y., Spachos, P., Hatzinakos, D., \& Plataniotis, K. N. Feasibility study of stress detection with machine learning through eda from wearable devices. In {\em ICC 2022-IEEE International Conference on Communications} {\bf 2022}, 4800--4805.

\bibitem[50]{zhu2022multimodal}
Zhu, L., Spachos, P., \& Gregori, S. Multimodal physiological signals and machine learning for stress detection by wearable devices. In {\em 2022 IEEE International Symposium on Medical Measurements and Applications (MeMeA)} {\bf 2022}, 1--6.

\bibitem[51]{jeon2020stress}
Jeon, T., Bae, H., Lee, Y., Jang, S., \& Lee, S. Stress recognition using face images and facial landmarks. In {\em 2020 International Conference on Electronics, Information, and Communication (ICEIC)} {\bf 2020}, 1--3.

\bibitem[52]{zhang2020video}
Zhang, H., Feng, L., Li, N., Jin, Z., \& Cao, L. Video-based stress detection through deep learning. {\em Sensors} {\bf 2020}, {\em 20}(19), 5552.

\bibitem[53]{cardone2020driver}
Cardone, D., Perpetuini, D., Filippini, C., Spadolini, E., Mancini, L., Chiarelli, A. M., \& Merla, A. Driver stress state evaluation by means of thermal imaging: A supervised machine learning approach based on ECG signal. {\em Applied Sciences} {\bf 2020}, {\em 10}(16), 5673.


\bibitem[54]{khoshkhahtinat2023multi}
Khoshkhahtinat, A., Zafari, A., Mehta, P. M., Akyash, M., Kashiani, H., \& Nasrabadi, N. M. Multi-Context Dual Hyper-Prior Neural Image Compression. {\em arXiv preprint arXiv:2309.10799} {\bf 2023}.

\bibitem[55]{kopaczka2019modular}
Kopaczka, M., Breuer, L., Schock, J., \& Merhof, D. A modular system for detection, tracking and analysis of human faces in thermal infrared recordings. {\em Sensors} {\bf 2019}, {\em 19}(19), 4135.

\bibitem[56)]{seo2022deep}
Seo, W., Kim, N., Park, C., \& Park, S. M. Deep Learning Approach for Detecting Work-Related Stress Using Multimodal Signals. {\em IEEE Sensors Journal} {\bf 2022}, {\em 22}(12), 11892--11902.

\bibitem[57]{mou2021driver}
Mou, L., Zhou, C., Zhao, P., Nakisa, B., Rastgoo, M. N., Jain, R., \& Gao, W. Driver stress detection via multimodal fusion using attention-based CNN-LSTM. {\em Expert Systems with Applications} {\bf 2021}, {\em 173}, 114693.


\bibitem[58]{walambe2021employing}
Walambe, R., Nayak, P., Bhardwaj, A., \& Kotecha, K. Employing multimodal machine learning for stress detection. {\em Journal of Healthcare Engineering} {\bf 2021}, 1--12.

\bibitem[59]{naegelin2023interpretable}
Naegelin, M., Weibel, R. P., Kerr, J. I., Schinazi, V. R., La Marca, R., von Wangenheim, F., Hoelscher, C., \& Ferrario, A. An interpretable machine learning approach to multimodal stress detection in a simulated office environment. {\em Journal of Biomedical Informatics} {\bf 2023}, {\em 139}, 104299.

\bibitem[60]{bailey2017electrodermal}
Bailey, R. L. Electrodermal activity (EDA). {\em The international encyclopedia of communication research methods} {\bf 2017}, 1--15.

\bibitem[61]{koelstra2011deap}
Koelstra, S., Muhl, C., Soleymani, M., Lee, J. S., Yazdani, A., Ebrahimi, T., Pun, T., Nijholt, A., \& Patras, I. Deap: A database for emotion analysis; using physiological signals. {\em IEEE transactions on affective computing} {\bf 2011}, {\em 3}(1), 18--31.

\bibitem[62]{breiman2001random}
Breiman, L. Random forests. {\em Machine learning} {\bf 2001}, {\em 45}, 5--32.
\bibitem[63]{cohen2009pearson}
Cohen, I., Huang, Y., Chen, J., Benesty, J., Benesty, J., Chen, J., Huang, Y., \& Cohen, I. Pearson correlation coefficient. In {\em Noise reduction in speech processing} {\bf 2009}, 1--4.

\bibitem[64]{myers2004spearman}
Myers, L., \& Sirois, M. J. Spearman correlation coefficients, differences between. {\em Encyclopedia of statistical sciences} {\bf 2004}, {\em 12}.

\bibitem[65]{tibshirani1996regression}
Tibshirani, R. Regression shrinkage and selection via the lasso. {\em Journal of the Royal Statistical Society Series B: Statistical Methodology} {\bf 1996}, {\em 58}(1), 267--288.

\bibitem[66]{guyon2003introduction}
Guyon, I., \& Elisseeff, A. An introduction to variable and feature selection. {\em Journal of machine learning research} {\bf 2003}, {\em 3}(Mar), 1157--1182.

\bibitem[67]{guyon2002gene}
Guyon, I., Weston, J., Barnhill, S., \& Vapnik, V. Gene selection for cancer classification using support vector machines. {\em Machine learning} {\bf 2002}, {\em 46}, 389--422.

\bibitem[68]{berger2015stress}
Berger, R. Stress, trauma, and posttraumatic growth: Social context, environment, and identities. {\em Routledge} {\bf 2015}.

\bibitem[69]{lundberg2020local}
Lundberg, S. M., Erion, G., Chen, H., DeGrave, A., Prutkin, J. M., Nair, B., Katz, R., Himmelfarb, J., Bansal, N., \& Lee, S. I. From local explanations to global understanding with explainable AI for trees. {\em Nature machine intelligence} {\bf 2020}, {\em 2}(1), 56--67.

\end{thebibliography}
\end{document}